\def\year{2024}\relax
\documentclass[letterpaper]{article} % DO NOT CHANGE THISW
\usepackage[submission]{aaai24}  % DO NOT CHANGE THIS
\usepackage{times}  % DO NOT CHANGE THIS
\usepackage{helvet}  % DO NOT CHANGE THIS
\usepackage{courier}  % DO NOT CHANGE THIS
\usepackage[hyphens]{url}  % DO NOT CHANGE THIS
\usepackage{graphicx} % DO NOT CHANGE THIS
\usepackage{natbib}  % DO NOT CHANGE THIS AND DO NOT ADD ANY OPTIONS TO IT
\usepackage{caption} % DO NOT CHANGE THIS AND DO NOT ADD ANY OPTIONS TO IT
\DeclareCaptionStyle{ruled}{labelfont=normalfont,labelsep=colon,strut=off} % DO NOT CHANGE THIS
\usepackage{algorithm}
\usepackage{algorithmic}
\usepackage{booktabs}
\usepackage{newfloat}
\usepackage{listings}
\floatstyle{ruled}
\newfloat{listing}{tb}{lst}{}
\floatname{listing}{Listing}
\usepackage{amsmath}

\usepackage{color}
\usepackage{xcolor}
\newcount\Comments  % 0 suppresses notes to selves in text
\long\def\NS#1{{\ifnum\Comments=1\color{red} [NS: #1]\fi}}
\long\def\JS#1{{\ifnum\Comments=1\color{purple} [JS: #1]\fi}}
\long\def\IS#1{{\ifnum\Comments=1\color{blue} [IS: #1]\fi}}

\title{Set-Based Retrograde Analysis:\\
Precomputing the Solution to 24-card Bridge Double Dummy Deals}
\author {
    Isaac Stone,\textsuperscript{\rm 1}
    Nathan R. Sturtevant, \textsuperscript{\rm 1,2}
    Jonathan Schaeffer \textsuperscript{\rm 1}
}
\affiliations {
    \textsuperscript{\rm 1} University of Alberta, Department of Computing Science\\
    \textsuperscript{\rm 2} Alberta Machine Intelligence Institute\\
    istone@ualberta.ca, nathanst@ualberta.ca jonathan@ualberta.ca, 
}

\usepackage{bibentry}
\begin{document}

\maketitle

\begin{abstract}
%Designing an algorithm to play the game of Bridge at a high level is a monumental task.

Retrograde analysis is used in game-playing programs to solve states at the end of a game, working backwards toward the start of the game. The algorithm iterates through and computes the perfect-play value for as many states as resources allow. We introduce \emph{setrograde analysis} which achieves the same results by operating on \emph{sets} of states that have the same game value. The algorithm is demonstrated by computing exact solutions for Bridge double dummy card-play. For deals with 24 cards remaining to be played ($10^{27}$ states, which can be reduced to $10^{15}$ states using preexisting techniques), we strongly solve all deals. The setrograde algorithm performs a factor of $10^3$ fewer search operations than a standard retrograde algorithm, producing a database with a factor of $10^4$ fewer entries. For applicable domains, this allows retrograde searching to reach unprecedented search depths.

% ORIGINAL ABSTRACT For the game of Bridge, current state-of-the-art programs rely on fast Double Dummy (DD) solvers to provide a useful approximation of . Using retrograde analysis that propagates sets of states instead of individual states, we present new algorithms for generating exact solutions for DD . For hands with 24 cards remaining to be played ($10^{27}$ scenarios) we strongly solve all deals. The program considers \textbf{more than $10^4$ fewer states} than than a standard retrograde algorithm would, and produces a database of approximately \textbf{3 orders of magnitude} smaller than produced by standard algorithms.
\end{abstract}

\section{Introduction}

%\ORG{Guidlines
%\begin{itemize}
%\item Use deal instead of hand for states
%\item Use {\em consistent set} instead of equivalent
%\item Use generate (as a verb) instead of create
%\item Avoid faster/smaller in evaluation
%\item Heuristic - avoid this word, it is ambiguous\end{itemize}
%}

% \ORG{Important point to go somewhere, even possibly introduction. Endgame databases normally give us a small computation which is a huge gain in search at the end. Card games don't converge in the way that Checkers does. So, it doesn't make sense to directly build endgame databases. \\
% 1. What is the problem (endgame databases and their use) \\
% 2. What is wrong (can't use in card games or Chinese Checkers) \\
% 3. How do we fix it? (set based search) \\
% 4. Summary of contributions. (We do it in Bridge)
% }

% start with bulleted contributions
Some of the early high-performance game-playing programs relied on retrograde analysis and endgame databases for strong play. The most notable example is Checkers, where 39 trillion endgame positions,
all those with 10 or fewer pieces, were used as part of the \textsc{Chinook} program \cite{Chinook}, and for solving Checkers \cite{Schaefer2007CheckersisSolved}. Endgame databases are also used widely in Chess programs \cite{ChessTablebases}, as well as in many other games (e.g., for solving Awari \cite{Awari}).

Endgame databases are most effective in games where there are far fewer positions at the end of the game than elsewhere. As a result, they have not been applied in games that do not have this property.
For instance, \citeauthor{sturtevant-thesis} (\citeyear{sturtevant-thesis}) noted that in 3-player Chinese Checkers a winning arrangement of a single player's pieces in the game has approximately $10^{23}$ possible permutations of the other player's pieces, making it infeasible to store all the variations of even a single winning configuration. While in Chinese Checkers each player has a unique endgame configuration (the other side's piece locations are irrelevant), in Go the locations of both side's pieces in a terminal state are important. Hence these games require significantly different analysis  \cite{berlekamp1994mathematical}.
%\NS{May want to remove Go example or move it elsewhere, as it points to similar abstraction as we are doing in this paper.}
In a 4-player trick-based card game such as Bridge, the last two tricks have $\binom{52}{2}\binom{50}{2}\binom{48}{2}\binom{46}{2}
= 1.9\times 10^{12}$ possible deals of the cards.
%($7.6\times 10^{16}$ for three tricks).
However, there are only 16 ways for each deal to play out, meaning it is trivial to solve but storing all states (as  done in Checkers) is difficult.

% Building high-performance game-playing programs was one of the early goals of AI research. Results in this area have been big success stories for AI. Programs are super-human at popular games such as backgammon, Checkers, Chess, go, and poker. Further, AI algorithms have been used to solve games such as Checkers and heads-up  (two player) limit poker. \IS{resolved terminology, still need to do citations.  I will work on citations during "down" time when I've handed off the draft for edits.}

These numbers suggest it might be impractical to build an effective endgame database for Bridge with, say, 6 tricks to play ($10^{27}$ states). This statement is true under the assumption that every unique endgame state must be stored independently. The contribution of this paper is to show how to avoid this assumption by representing endgame states as sets. This idea, along with other symmetry reduction techniques, makes it feasible to use retrograde search to compute all 24-card (6-trick) Bridge double-dummy (DD) endgames in a week on appropriate hardware using just 50GiB of storage, something that was historically hard to imagine. %\JS{New last line}
% \JS{Drop the next sentance. Misleading}
% This translates to approximately $9\times10^{16}$ states per byte.

% Bridge is a popular card game that remains an open problem. The game is challenging because it has four players, involves random deals of 52 cards, has imperfect information, and an enormous search space. There has been limited AI success in handling the  phase, where the computer has to play cards in such a way as to win the maximum number of tricks. Algorithms, such as Partition Search \cite{ginsbergFirstPartition96},  enable faster Double Dummy (DD) solutions (solving a deal when all four hands are known), which in turn provide a strong approximation of . Even so, the best computer  does not match that of strong humans.

% For some games (e.g., solving Checkers \cite{Schaefer2007CheckersisSolved}) retrograde analysis has proved to a valuable tool. In this paper the principles of retrograde analysis are combined with the Partition Search idea of using sets of states to exactly solve Double Dummy deals. The result is a new set-based retrograde algorithm that achieves lossless generalization and significant compression. %The algorithms have been validated by strongly solving all $10^{15}$ DD problems with 24 or fewer cards remaining.
%\footnote{specifically, we have solved all states in which a multiple of 4 cards remain, we don't solve intermediate states}.
% TODO:

This paper describes our set-based approach to endgame databases, making the following contributions.

\begin{itemize}
    %\item As far as we know, this is the first application of Endgame Databases [JS: change to retrograde analysis; introduce endgame databases in the background section] to card games.  
    %\item We show that Endgame Databases can be extended to all 24 card Double Dummy problems using a set-search [JS: set-search is not defined].
    \item We present a new set-based retrograde analysis algorithm, \emph{setrograde analysis}, inspired by the ideas in Ginsberg's Partition Search \cite{ginsbergFirstPartition96}.
    Whereas standard retrograde analysis computes a value for every state, setrograde analysis generalizes a state into a set where all members have the same game-theoretic value. The algorithm can skip over many of the states that are subsumed by the set.  Replacing states with sets leads to a large degree of state-space compression, by mapping an exponentially growing state-space to a smaller set-space.
    %Setrograde analysis can be used to strongly solve large state spaces in less time and space than the previous state-of-the-art.
    %The algorithm is demonstrated with 24-card Bridge hands, and can be generalized to other game domains.
    %\IS{I'm wondering if it might be better to present this all as a single contribution.  i.e. first point (setrograde analysis) is our contribution, and then we combine the last sentence of the first point with the second and third bullets, which are really just paragraphs describing the original contribution, and pre-stating results to emphasize the effectiveness of the approach/significance of the contribution.}
    %\IS{this might be a good spot to reference the statement of infeasibility for multiplayer Chinese Checkers endgame databases from \cite{SturtevantPhdThesis02} page 119}
    %\item We present general algorithms for performing heuristic free set searches.
    %\item We perform our set-searches bottom-up using retrograde analysis.
      %\item \JS{To be discussed. I need to know more about this data structure.} We introduce a Graduated Tree data structure and pairwise merging algorithms to store and compress exact models of the world.
%This enables the rapid insertion of entries into the tree and the looking up of values in the solution database during construction. The latter is used to query the final solution to extract the exact value of a 24-card Bridge hand.

      \item The algorithm is demonstrated using 24-card Bridge deals. The set database contains 4 orders of magnitude (OOM) fewer sets than there are states in a traditionally generated database. The set database was constructed using 3 OOM less computing resources than would be needed for a traditional 24-card database. This enabled an 800 trillion state state-space to be solved in a week using a single multi-core machine.
      %\JS{Deleted: though the exact improvement is unclear since traditional methods could not solve beyond 16 cards on the resources available to us.} \JS{Still too imprecise} \IS{I'm not sure where to go from here, since there really is no clear baseline.  All we have is the size of the statespace compared with the size of the setspace, or the speed/database sizes of the practical zero-window retrograde solver vs the practical setrograde solver (but we don't have the sizes beyond 3 tricks for a practical standard database, and it's not heavily optimized)}

\end{itemize}

This work generalizes retrograde analysis, allowing it to have more impact in applicable domains. In particular, the reduced computing and storage needs mean that endgame database technology can be scaled to unprecedented levels. 

\section{Background and Related Work}
% \ORG{This section is just an overview (citations) of work on Bridge, applications of retrograde analysis, solved games, and search algorithms in games (perfect vs imperfect information, double dummy)}

\begin{figure}
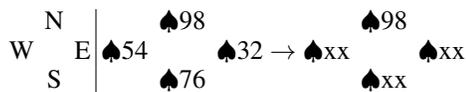

\begin{center}
\begin{tabular}{c c c | c c c c c}
&N&&&$\spadesuit$98&&$\spadesuit$98\\
W&&E&$\spadesuit$54&&$\spadesuit$32 $\rightarrow$ $\spadesuit$xx&&$\spadesuit$xx\\
&S&&&$\spadesuit$76&&$\spadesuit$xx\\
\end{tabular}\\
\end{center}
\caption{$\spadesuit$98 wins two tricks regardless of the locations of lower cards in the perfect-information deal (left) and set of deals (right); \emph{x} refers to any low card.
%Players are referenced by cardinal direction, with the North hand depicted at the top, South at bottom, East to the right, and West to the left. East is next-to-play.
Play proceeds clockwise starting with North (N) until each player has played one card (East, South, West). The highest card wins.  The winning player is next-to-play.}
\label{fig-x-example}
\end{figure}

Double-dummy problems are the perfect-information variant of Bridge card play. A deal of $n$ cards (with $n$ as a multiple of 4) has an exact integer evaluation under perfect play, ranging from 0 to $\frac{n}{4}$ --- the number of tricks won. See Fig.~\ref{fig-x-example} (left) for a one-suit, two-card deal; the four player around the table are referred to as North, South, East and West.

The idea behind retrograde analysis is to solve a game from the end towards the start. In the example of Bridge, this works by enumerating all deals where each player has one card (1 trick), solving them, and then storing the results. Then one can move backwards in the game to consider all deals where each player has two cards (2 tricks).
For a given 2-trick deal, taking the maximum result of all successor 1-trick deals (already computed) produces the correct result. Given sufficient computational and storage resources, one could continue to obtain the 3-trick results, and so on. Most often the algorithm is expressed as solving depth $d$ (where $d$ is the number of tricks in Bridge or the number of pieces on the board as in Chess) given the precomputed results from depth $d-1$.
The computational cost and storage requirements typically grow exponentially in $d$. There are numerous enhancements to the basic algorithm that can improve its performance \cite{sturtevant2017ccsolve}.
%To date, retrograde analysis has been applied to computing the value for all states $\leq{d}$, and storing those states (often called an endgame database or tablebase).
Retrograde analysis is typically used to compute and store the value for all states up to a given depth $d$, producing a comprehensive database (often called an endgame database or tablebase).

Since evaluations are discrete, $n$-card deals may be grouped into {\em consistent} sets, meaning that all states in the set share the same evaluation. %a bound on their evaluation.
% with identical evaluations, or sets with a bound on the evaluations\IS{could we define this as a "consistent" set here? — useful term to have for later}. 
%
Partition Search \cite{ginsbergFirstPartition96} is a forward minimax solver with a set-based backup heuristic that generates a consistent set at each state in the search.
% precisely generalize evaluations from states to sets. 
Ginsberg showed a branching factor reduction by using these sets as transposition table entries. DD solvers based on Partition Search remain a state-of-the-art tool for modern mechanical Bridge players.

%\NS{July 10: This is not detailed enough. Need to say what the diagram is. What the different sides represent. The fact that we are looking at double-dummy play, where all cards are face up. The fact that normal hands have four suits, but we just use one for simplicity. In this case each player plays a card in order and the highest card wins.} 
%\IS{Addressed in figure caption, with the exception of specifying one-suit Bridge.  I don't think the lack of specification will be confusing to anyone familiar with the game, and the addition of a specification would seem to require taking the reader out of the thread, the only reason I can think of to add that (maybe I'm missing something) is to justify the complexity of the game and emphasize that this is a simplification.}
Bridge deals are colloquially described by the number of cards in each suit, and the ranks of relevant high cards. A sufficiently low card, that does not affect the result, is usually referred to as \emph{x}. This notion is formalized in the set representation we use.  Any card denoted \emph{x} is interchangeable with any other \emph{x}, and is strictly lower than a ranked card. For example, Fig.~\ref{fig-x-example} (right) shows a set containing $6!/(2!)^3 = 90$ states that Partition Search might discover by generalizing the state in Fig.~\ref{fig-x-example} (left).

% set-based generalization of the left-hand side that represents 90 scenarios. The storage difference is significant: $6!/(2!)^3 = 90$ states versus 1 set.

Partition Search maintains consistent sets by heuristically backing up cards that can be proven to never win by rank.  Available DD solvers rely on expert heuristics for move ordering, set generation, and early search cutoffs. For retrograde searching, we present a more general algorithm for proving consistent sets and a framework in which the need for expert knowledge is eliminated.
%\NS{July 10: Key idea to communicate here: Partition search starts with a concrete hand and returns an abstraction. } \IS{not sure how to address at the moment.  Will clarify tomorrow.}

\setlength{\tabcolsep}{2pt}

There have been previous attempts at set-based search algorithms in games, but they bear little resemblance to our approach and they have not been applied (or are even  relevant) to backward search. Our algorithms are designed to strongly solve large state-spaces (finding the perfect play result for all reachable states) using retrograde analysis.  Previous approaches have aimed at weakly solving state-spaces using a top-down approach. Retrograde state-space search has seen success in board games including Checkers \cite{Schaefer2007CheckersisSolved}, Chinese Checkers \cite{ChineseCheckersStrongSolve2020}, and Chess \cite{ChessTablebases}. Retrograde analysis has been applied to a subset of Skat (null) endgames \cite{furtak-thesis}. All of the set-wise approaches that we know of, including Proof-Set Search \cite{ProofSetSearchMartinMuller}, Method of Analogies \cite{MethodOfAnalogiesAdelson-Velsky1988}, and even Partition Search \cite{ginsbergFirstPartition96}, have been designed for top-down search, and targeted toward weak solutions \cite{,haglundDD2014,Beling2017PsearchRevisited}. Set-based search has been applied successfully in planning, where sets of states are often represented using Binary Decision Diagrams \cite{edelkamp2015bdds}.

\section{Overview}

Here we provide a high-level overview of a set-based approach to retrograde analysis. The ideas are illustrated using examples from Bridge.

\subsection{State-Space Reduction}

This section is specific to the game of Bridge, but is important for illustrating the search-space reductions that are possible in Bridge and in set-based search.

A 24-card (6 tricks) endgame database contains the solution to all ${52 \choose 6} {46 \choose 6} {40 \choose 6} {34 \choose 6}= 10^{27}$ ways to distribute the cards from a standard deck.  Additionally, there are 5 trump suits to consider (clubs, diamonds, hearts, spades, and no trump). Hence there are $5 \times 10^{27}$ unique deals represented in the 24-card database.

A well-known optimization used in card-game transposition tables is to represent cards using relative ranks instead of absolute ranks \cite{haglundDD2014}.  If there are 8 spade cards in play, the lowest one is always represented as a 2, the next lowest as a 3, and so on. Thus, $13 \choose 8$ possible deals (1,287) are reduced to 1 representative deal. % since all of the deals have identical strategies.
This does not affect card-play mechanics, but it reduces the number of 24-card states by roughly 12 OOM.  Several minor symmetry-related optimizations (not discussed here) can be applied to remove approximately one OOM, leaving  $8\times10^{14}$ (800 trillion) distinct states that must be evaluated and stored in a 24-card endgame database.

%\IS{if you're wondering, the actual reduction from suit symmetries is about a factor of 24.  4 suits can be the "first" suit; 3 options for second suit, 2 for third suit.  4!.  It's not exact since there are some other symmetries that have to be accounted for but minor ones.}
%\IS{This paragraph (and table?) through \textbf{Set-Space Reduction} may belong in results, or at the end of \textbf{Set-Space Reduction}}

Table \ref{fig:stateSpaceSize} provides the number of states given the number of cards remaining.
\emph{Upper Bound} is the number of states a state-based retrograde solver would need to both solve and store (after the rank and symmetry state-space reductions).
\emph{Setrograde} is the number of \emph{sets} stored in our set-based database through 24 cards. At the time of this writing, the 28 card results are being computed.
%Our algorithm is working on generating the 28-card database, so numbers for 28-cards and beyond are omitted.
\emph{Lower Bound} is a bound on the number of sets in a complete DD database, assuming an approach that partitions the state-space by the distribution of suits between the players. The data in this table is discussed in more depth later in the paper.
%As long as partitions are maintained between shapes, the database must contain at a minimum, one set for each shape that is reachable from a 52-card starting position.
%\JS{Shape ommitted as it is not used anywhere else.}

Table \ref{fig:stateSpaceSize} demonstrates two major state-space reductions, and their scaling. As the number of cards increases, the relative-rank state-space reduction increases rapidly at first, peaking with a 12 OOM reduction at 24 and 28 cards.  Beyond 28 cards, this reduction decreases with depth. Once 52 cards remain, all 13 cards in each suit are represented at all times, and no rank reduction exists. (The reduction that remains is due to symmetries.)  The size of the set-space considered by the setrograde analysis algorithm (introduced in this paper) are also shown. Set-space growth seems to be decelerating (e.g., growing by almost 2 OOM going from 16 to 20 cards, but only 1 OOM in going from 20 to 24), and grows much more slowly than the state-space. As the number of cards increase, larger sets are produced.
%While the upper bound continues to increase rapidly, the lower bound actually inverts at 40 cards remaining. \IS{@Jonathan or @Nathan: this may be a good place to clearly word the claims we want to make.}

% We also provide the number of distinct ``shapes" (ways of distributing the suits between the hands, along with the number of reachable shapes.  From 16 to 48 cards remaining, there are shapes that cannot be legally reached in play and, therefore, can be omitted from the databases. 
% \ORG{Explain upper bound, lower bound, and setrograde.}
% \JS{Do we use shapes anywhere? It does not affect the algorithms. Delete?}\IS{We never fully explain them.  The reason I include them, with this \emph{brief} description, is that shapes are a metric on what needs to be solved at each level (a lower bound on the number of sets in fact, given the representation we use).  At size 6 we're only about order of magnitude away from the number of shapes in the largest sizes.  And we have as many 24-card shapes as 52 card shapes.  I have no strong objections to removing shapes.  The reason for suggesting the inclusion is that it can re-appear in results to show that database size may curve over, and shapes provide the most optimistic view of scaling.  If that's an argument we want, we should keep it.  If it's too difficult or distracting to make that argument, then we shouldn't} \NS{If shapes stay in, we need to make it clear that shapes are large, so the task is much harder.}

\begin{table}[t]
\small
\setlength{\tabcolsep}{3.5pt}
\begin{center}
\begin{tabular}{ c c c c c }
%\begin{tabular}{ c r r r r }
\toprule
Cards&\multicolumn{1}{c}{States}&\multicolumn{1}{c}{Upper Bound}&\multicolumn{1}{c}{Setrograde} &\multicolumn{1}{c}{Lower Bound}\\
\midrule
4 &$3\times10^{7}$&$2\times10^{2}$&$7\times10^{1}$&$2\times10^{1}$\\
8 &$9\times10^{12}$&$8\times10^{4}$&$8\times10^{3}$&$5\times10^{2}$\\
12&$4\times10^{17}$&$3\times10^{7}$&$5\times10^{5}$&$7\times10^{3}$\\
16&$3\times10^{21}$&$1\times10^{10}$&$3\times10^{7}$&$6\times10^{4}$\\
20&$7\times10^{24}$&$3\times10^{12}$&$1\times10^{9}$&$4\times10^{5}$\\
24&$5\times10^{27}$&$8\times10^{14}$&$2\times10^{10}$&$2\times10^{6}$\\
28&$1\times10^{30}$&$2\times10^{17}$&-&$7\times10^{6}$\\
32&$6\times10^{31}$&$3\times10^{19}$&-&$2\times10^{7}$\\
36&$1\times10^{33}$&$4\times10^{21}$&-&$4\times10^{7}$\\
40&$5\times10^{33}$&$4\times10^{23}$&-&$4\times10^{7}$\\
44&$4\times10^{33}$&$4\times10^{25}$&-&$3\times10^{7}$\\
48&$3\times10^{32}$&$2\times10^{27}$&-&$9\times10^{6}$\\
52&$3\times10^{29}$&$1\times10^{28}$&-&$2\times10^{6}$\\
% %%
% Cards&States&Reduced&Reduction&Shapes&Reachable\\
% \midrule
% 4 &$3\times10^{7}$&$2\times10^{2}$&$10^5$&$2\times10^{1}$&$2\times10^{1}$\\
% 8 &$9\times10^{12}$&$8\times10^{4}$&$10^8$&$5\times10^{2}$&$5\times10^{2}$\\
% 12&$4\times10^{17}$&$3\times10^{7}$&$10^{10}$&$7\times10^{3}$&$7\times10^{3}$\\
% 16&$3\times10^{21}$&$1\times10^{10}$&$10^{11}$&$6\times10^{4}$&$6\times10^{4}$\\
% 20&$7\times10^{24}$&$3\times10^{12}$&$10^{12}$&$4\times10^{5}$&$4\times10^{5}$\\
% 24&$5\times10^{27}$&$8\times10^{14}$&$10^{13}$&$2\times10^{6}$&$2\times10^{6}$\\
% 28&$1\times10^{30}$&$2\times10^{17}$&$10^{13}$&$7\times10^{6}$&$7\times10^{6}$\\
% 32&$6\times10^{31}$&$3\times10^{19}$&$10^{12}$&$2\times10^{7}$&$2\times10^{7}$\\
% 36&$1\times10^{33}$&$4\times10^{21}$&$10^{12}$&$4\times10^{7}$&$4\times10^{7}$\\
% 40&$5\times10^{33}$&$4\times10^{23}$&$10^{10}$&$6\times10^{7}$&$4\times10^{7}$\\
% 44&$4\times10^{33}$&$4\times10^{25}$&$10^{8}$&$6\times10^{7}$&$3\times10^{7}$\\
% 48&$3\times10^{32}$&$2\times10^{27}$&$10^5$&$3\times10^{7}$&$9\times10^{6}$\\
% 52&$3\times10^{29}$&$1\times10^{28}$&$10^1$&$2\times10^{6}$&$2\times10^{6}$\\
% %%
\bottomrule
\end{tabular}
\caption{State-Space Size and Reductions for Bridge}
\label{fig:stateSpaceSize}
\end{center}
\end{table}
% \IS{(TABLE: remove reachable, switch titles to be consistent, use Lower bound.  ALSO: make the case for the two reductions branching.  Change from reduction to where are we, reduced is upper bound, shapes is LB.}

\subsection{Set-Space Reduction}
Retrograde analysis, shown in Alg. \ref{alg:Setrograde0}, works by iterating over all states at depth $d$ (lines 3-4, 12-13), computing a state's value based on the successor states at depth $d-1$ (line 5), and storing it into the database (line 7).
%This usually has to be done repeatedly until all the  states have their value resolved.
%Each state's value would then be saved. 
Applying the same approach to 24-card Bridge would require repeatedly iterating over almost $10^{15}$ states (assuming the reductions given above), requiring storage of $\sim 10^{15}$ bytes (depending on possible compression approaches).

Our set-based algorithm, \emph{setrograde analysis}, generates a database of consistent sets. 
We start with a simpler version of the algorithm, also shown in Alg \ref{alg:Setrograde0}, before refining our description into a more efficient version.
% \IS{creating and updating is a bit confusing.  I usually say generating, but I get what the goal is here.  Maybe: "Our set-based algorithm, \emph{setrograde analysis}, instead generates a database of sets." (simple seems fine here)}
As with retrograde analysis, the simpler implementation works by iterating through all states at depth $d$ (lines 3-4, 12-13).
Each state is queried in the database to see if it matches any of the sets that have already been computed (line 5). If such a set is found, the state's value is known and the algorithm moves on to the next state. Otherwise, a search routine finds a consistent set containing the new state (line 9), and that set is added to the database (line 10).
% \IS{NEW (potential) LAST SENTENCE: "Otherwise, a search routine finds a consistent set containing the new state, and we add the set to the database."} \IS{previous version commented out here.  The details that are removed re-appear later anyway.}%If not, then generalized versions of the state are created, and one is chosen to add to the database (usually the most general one).

We highlight here some of the challenges in creating a fast setrograde algorithm:

\begin{enumerate}
\item Generalization.
This additional routine (line 9) is performed at every state. The cost of this operation must be less than the cost of performing the retrograde operation for each of the states encompassed by the set returned.
\item Querying. It is more complex to look for a state in a database of {\em sets} than to look for a state in a database of {\em states} (line 5). $databaseLookup$ performs multiple queries at each state. Without a fast and scalable implementation, queries become a computational bottleneck.%\IS{Again, inefficient implementations can negate all gains — I don't think this point is clear here, and I think this is the point we are looking for.}
\item Iteration. Iterating over all members of a large state-space will be expensive even if the cost per state is small.
To scale computation, we must be able to iterate through the set-space without considering each state in the (much larger) state-space.
\end{enumerate}

\begin{algorithm}[t]
\caption{Retro/Setrograde Analysis}
\label{alg:Setrograde0} 
    \begin{algorithmic}[1]
    \FOR{$d\gets 1..D$}
        \STATE $EDB_d \gets \emptyset$
        \STATE {$s \gets \textit{firstState}(d)$}
        \WHILE{ $s \neq null$}
            \STATE $v \gets databaseLookup(s,d)$
            \IF{$\textit{algorithm} =\textit{retroGrade}$}
            
                \STATE $EDB_d[s] \gets v$
            \ELSIF {$EDB_d[s]=undefined$}%$v = undefined$}
                \STATE $t \gets \textit{generalizeToSet}(s,v)$
                \STATE $EDB_d[t]\gets v$
            \ENDIF
            \STATE $s \gets nextState(s,d)$
        \ENDWHILE
    \ENDFOR
\end{algorithmic}
\end{algorithm}

Each of these are briefly described. Generalization and iteration are then illustrated using Bridge.

%First, consider the generalization issue.
\subsubsection{Generalization}
In retrograde analysis, if a state is reached that is not in the database, then it will have its value computed and added to the database. Instead, setrograde analysis finds a generalization of the state (a set) in which all states have the same value (consistent).
It does this using a generate-and-test approach, where sets are generated until the best consistent set (along some metric) is found. In Bridge, this is done by replacing some cards with \emph{x}'s.

Generalization is illustrated in Bridge in Section \ref{sec:generalization}.
%The generalization function can be expensive.
If all possible ways of generalizing a state are considered, this function would be very expensive.
% \IS{this part is a bit inexact I think.  First, we should use sets instead of abstractions, second "all possible" sets requires (difficult) contextualization since "all possible" depends on representational format, or else is the powerset of the statespace (aka, big).  I'd suggest we focus on one specific bad implementation, and describe it more concretely.  Or simply emphasize that we would like to evaluate as few sets as possible (Note that while in the naive method, we would end up doing a lot of lookups, we would actually generalize the same number of sets.  It's just that duplicate detection becomes the computational bottleneck.  Hopefully, this clarifies things a bit)}
%\IS{we provide rudimentary details for Bridge in section 5, and an in-depth explanation with examples in the technical appendix}
In the Technical Appendix, we show how the cost can be reduced by considering only a subset of generalizations.

\subsubsection{Querying} 
In retrograde analysis, a ranking or perfect hash function is used to map every state to a unique number. States are not explicitly stored; the offset in memory uniquely identifies a state.
% and indexing into the database by that number retrieves the value for that state.
In setrograde analysis, sets must be explicitly stored. Thus,
looking up a state in the database (line 5) involves matching a state to a set, a more complicated operation.
Querying can be done efficiently using an appropriate data structure. One such data structure is briefly discussed in the Technical Appendix.

\subsubsection{Iteration} Some states can be bypassed, given that they are members of sets already in the database. 
Similar to how backjumping is used in DPLL search \cite{gaschnig1979performance} to avoid searching irrelevant portions of a tree, it is possible to identify portions of the space that have already been solved, and jump past them. The exact approach is dependent on the state representation being used. Low-cost iteration is illustrated in Bridge in Section \ref{sec:iteration}.

\subsection{Example 1: Generalization in Bridge}\label{sec:generalization}

In the following deal, East is on lead.  North and South will take 2 tricks with the two highest spades:

\setlength{\tabcolsep}{2pt}
\begin{center}
\begin{tabular}{c c c c | c c c c}
&$\spadesuit$98&&&&&N\\
$\spadesuit$32&&$\spadesuit$54&&&W&&E\\
&$\spadesuit$76&&&&&S\\
\end{tabular}
\end{center}

\begin{figure}[t]
\setlength{\tabcolsep}{2pt}
\begin{center}
\begin{tabular}{c c c | c c c | c c c}
a)&$\spadesuit$98&&b)&$\spadesuit$98&&c)&$\spadesuit$98&\\
$\spadesuit$3x&&$\spadesuit$54&$\spadesuit$xx&&$\spadesuit$54&$\spadesuit$xx &&$\spadesuit$5x\\
&$\spadesuit$76&&&$\spadesuit$76&&&$\spadesuit$76\\
\hline
d)&$\spadesuit$98&&e)&$\spadesuit$98&&f)&$\spadesuit$98&\\
$\spadesuit$xx&&$\spadesuit$xx&$\spadesuit$xx&&$\spadesuit$xx&$\spadesuit$xx&&$\spadesuit$xx\\
&$\spadesuit$76&&&$\spadesuit$7x&&&$\spadesuit$xx\\
\hline
g)&$\spadesuit$9x&&h)&$\spadesuit$xx&&&N\\
$\spadesuit$xx&&$\spadesuit$xx&$\spadesuit$xx&&$\spadesuit$xx&W&&E\\
&$\spadesuit$xx&&&$\spadesuit$xx&&&S\\
\end{tabular}
\caption{Sets generated by replacing 0 or more low-rank cards with \emph{x}'s. East is on the lead.}
\label{fig:generalize}
\end{center}
\end{figure}

To generalize this deal we must produce candidate sets. Here we use the 8 candidates (a-h) shown in Fig.~\ref{fig:generalize} that were generated by replacing one or more low-rank cards with \emph{x}'s. 
It can quickly be verified that sets a-f contain only states in which North and South take 2 tricks. %(As in the original state, North and South win with both the 9$\spadesuit$ and 8$\spadesuit$.)
For example, consider verifying set f using minimax search. East, South, and West each have one legal move (play an \emph{x}), after which North plays either the 9 or the 8, winning the trick. The resultant set using relative card ranks is:

\begin{center}
\begin{tabular}{c}
\begin{lstlisting}
   $\spadesuit$4
$\spadesuit$x     $\spadesuit$x
   $\spadesuit$x
\end{lstlisting}
\end{tabular}
\end{center}

% \JS{Isaac, check the text from here to the end of the subsection  for correctness.}\IS{Looks good (removing comments)}

Having taken one trick, North and South must take one more trick in this resultant set. A search for this set in the 4-card database returns a value of one trick for North and South. Therefore this candidate set is a valid generalization. Note that in general, the exact set being looked for may not be in the database. Instead the overlap of several entries may be equivalent to the set looked for. The value of the overlapped sets would be the minimum of their values.

%A scan through the 4-card database fails to find a set sharing a deal with the resultant set, in which North and South do not take one trick.  In other words, North is guaranteed to take one trick exactly in any state in the resultant set.  Therefore this candidate set is a valid generalization.

Sets g and h each contain at least one deal in which North and South take only one trick, for instance:

\begin{center}
\begin{tabular}{c}
\begin{lstlisting}
    $\spadesuit$96
$\spadesuit$87      $\spadesuit$54
    $\spadesuit$32
\end{lstlisting}
\end{tabular}
\end{center}

%A similar analysis occurs for set g.
Consider trying to verify the correctness of g. Similar to the analysis of f above, East, South, and West again have one legal move and North has 2. North can play an \emph{x} in which case North may fail to win the current trick.  While a complex analysis can be performed when it is unclear which card wins a trick, it is sufficient here to note that in any case where North and South fail to win this trick, North and South will be unable to take 2 total tricks. Therefore, if North plays an \emph{x} on best play, set g must not be consistent.  North therefore plays the 9, producing the following resultant set:

\begin{center}
\begin{tabular}{c}
\begin{lstlisting}
   $\spadesuit$x
$\spadesuit$x     $\spadesuit$x
   $\spadesuit$x
\end{lstlisting}
\end{tabular}
\end{center}

North and South must take one more trick in this resultant set. Searching the 4-card database finds the following entry, which overlaps the set being looked for. Here North and South take 0 more tricks:

\begin{center}
\begin{tabular}{c}
\begin{lstlisting}
   $\spadesuit$x
$\spadesuit$4     $\spadesuit$x
   $\spadesuit$x
\end{lstlisting}
\end{tabular}
\end{center}

This means on North's action (the $\spadesuit$9) set g is not consistent.  Since North has no legal actions that leads to consistent resultant positions, set g is not consistent and cannot be added to the database.

\subsection{Example 2: Low-Cost Iteration in Bridge}
\label{sec:iteration}

To reduce computation, we skip over states that are members of the set just added to the database. We only solve and generalize \emph{independent} states — states not yet represented in the database. Each time a set is added to the database, we generate one or more potential next-states. These states are constructed by applying minor changes to the previous set that give rise to states that are provably not represented by that set.
The next states are placed into an open list (as in the A* algorithm \cite{Astar}) for further consideration.
From the previous example, if we add to the database the set:

\setlength{\tabcolsep}{2pt}
\begin{center}
\begin{tabular}{c c c c | c c c c}
&$\spadesuit$98&&&&&N\\
$\spadesuit$xx&&$\spadesuit$xx&&&W&&E\\
&$\spadesuit$xx&&&&&S\\
\end{tabular}
\end{center}

\noindent
it would be redundant to evaluate any other state in which North holds $\spadesuit98$. Therefore an independent state can be produced efficiently by trading the $\spadesuit8$ for one of the $\spadesuit$x cards.  For example, we could produce the following set (one of three such possibilities):

\begin{center}
\begin{tabular}{c}
\begin{lstlisting}
    $\spadesuit$9x
$\spadesuit$xx      $\spadesuit$xx
    $\spadesuit$8x
\end{lstlisting}
\end{tabular}
\end{center}

\noindent
If this set is consistent then it will be turned into a state by replacing the \emph{x}'s with low values (one of 720 possibilities):

\begin{center}
\begin{tabular}{c}
\begin{lstlisting}
    $\spadesuit$97
$\spadesuit$32      $\spadesuit$54
    $\spadesuit$86
\end{lstlisting}
\end{tabular}
\end{center}

\noindent
This new state goes onto the open list and eventually gets generalized using the process shown in Example 1. This process repeats until all deals in the state-space are represented by a set in the database.  Through this process, a setrograde solver can consider far fewer states than it would with the standard iteration over all states, reducing computation costs.  Further insight into this process is included in the technical appendix.%\IS{I modified this slightly.  I'm not sure what more to put here without it exploding into intricacy.  I think this is a good place to point toward the technical appendix.}

\section{Setrograde Analysis}

%\NS{I suggest covering the representation used here.}
%\NS{How do you map a state to a set? You have to find the most general set. Talk about this briefly.}
%\NS{Partition search has to build sets at runtime, while we do it offline. }

Now we turn to a more complete description of setrograde analysis, identify further bottlenecks, and describe how these are implemented efficiently.

These  definitions are used to describe setrograde analysis:
\begin{itemize}
    \item $D$: maximum retrograde distance — in Bridge, number of tricks
    \item $d$: retrograde distance $d\in1..D$ (that is, distance to terminal state)
    %\item $\mathbf{S}$: all states in the state space %$\forall d \in 1. .D$
    \item $\mathbf{S}_d$: all states at distance $d$ (from terminal state) %\NS{I think we should avoid calling this a set, even if it is, to avoid confusion}
    % \item $s_{d,i}$: the $i^{th}$ state in $\mathbf{S}_d$ \JS{verify usage}\IS{We can remove the subscripts from alg 3 and it will still be correct. So we can remove this}
    %\item $\mathbf{T}$: a set of sets of states \JS{Not used}\IS{I believe that is correct}
    \item $\mathbf{T}_d$: a set of sets of states such that all encapsulated states have a retrograde distance $d$
    % \item $\mathbf{t}_{d,i}$: the $i^{th}$ set of states at in $\mathbf{T}_d$
\end{itemize}

\begin{algorithm}[t]
\caption{General Setrograde Analysis}
\label{alg:Setrograde2} 
    \begin{algorithmic}[1] 
    \FOR{$d\gets 1..D$}
         \STATE $\mathbf{T}_d \gets \emptyset$
         \STATE $EDB_d \gets \emptyset$

        \STATE $s \gets \textit{nextIndependentState}(\mathbf{T}_d)$
        
        \WHILE{$s\neq$ \textbf{null}}

            %\IF {$EDB_d[s] = undefined$}
            \STATE $v \gets \textit{databaseLookup}(s,d)$

            \STATE $t \gets \textit{generalizeToSet}(s,v)$

            \STATE $\mathbf{T}_d \gets \mathbf{T}_d \cup t$
                
            \STATE $EDB_d[t]\gets v$
            %\ENDIF

                \STATE $s \gets \textit{nextIndependentState}(\mathbf{T}_d)$
    
        \ENDWHILE
        \STATE \textit{compactEDB}($EDB_d$)
    \ENDFOR
\end{algorithmic}
\end{algorithm}

Setrograde analysis (Alg. \ref{alg:Setrograde2}) parallels its state-wise predecessor with three key modifications. First, a set $\textbf{T}_d$ is maintained to track solved states (which are not stored explicitly but found in the union of all sets in $\textbf{T}_d$). %\JS{But definition says it contains sets, not states}\IS{it contains sets, but the union of the sets is the solved states, and we are looking for states that are not in the union of T}
%\IS{commented out old version}
%, allowing the algorithm to skip solved states (they are already covered by a set in the database)\IS{remove parenthetical}.
%In the state-wise formulation, this was not neccesary, since iterating through all states does not cause duplication. 
Second, at each iteration, a state $s\in\mathbf{S}_d|s\notin \mathbf{T}_d$ is evaluated — that is, states that are already solved are not re-evaluated.  While re-evaluation does not occur in the state-wise formulation, it could occur in the set-wise formulation if the states were evaluated iteratively without validating independence.  Finally, each solved state is generalized to a consisistent set $t$. 
Set $t$ is stored in the database instead of the individual state.
%\JS{One instance in $t$ not all sets in $t$, correct?}\IS{lower case t is just a set, not a set of sets.  so yes, but more accurately there is only one instance of t}

% \NS{This is covered above.} 
Alg.~\ref{alg:setrogradeHelpers} provides the helper functions needed for Alg.~\ref{alg:Setrograde2}.
The generalization process from a state to a consistent set can be done in many ways. One method is a backup heuristic, such as the one used in most modern Double Dummy solvers \cite{ginsbergFirstPartition96}.
The heuristic produces a single consistent set $\emph{t}$.
%\NS{Could just better justification}
This approach opts for simplicity, ignoring generality at each step and incurring a compounding performance penalty in both computation and storage. For our setrograde solver, we use a generate-and-test approach.
%For Bridge, we implement this using a binary search, and may apply to other games as well.
% , which is guaranteed to have the desired properties (that is, to contain the state we have evaluated and to contain only evaluation-equivalent states). 
Multiple candidate sets are produced, which may or may not be consistent.
Each is evaluated for consistency. Any metric (for instance $|\emph{}{t}|$) can be used to select which consistent set to add to the database.

Setrograde analysis is not guaranteed to produce the smallest possible database by any metric.  The order in which states are evaluated can affect which sets are added to the database.  The metric used to select which consistent candidate set is added to the database (or indeed whether multiple sets are added) can affect the composition of the final database.  Some metrics (including $|\emph{t}|$) can result in ties, and the tiebreak can affect the composition and size of the database.  Anything that affects the size of the database may also affect computation time since speed is positively correlated with the number of states that are evaluated and generalized.  Future work might establish stronger performance guarantees or tighter bounds.

The lower bound on the number of sets required for a database is based on an ideal set representation in which all deals with the same game value are represented in a single set.  In that case, we need exactly one set for each game value.  
% At a retrograde depth $d$, North and South may take $v\in0..d$ tricks, so there are d+1 possible game values. This is a loose bound, and our implementation lends itself to a tighter bound.  
In Bridge we partition the state-space based on the distribution of suits between the four hands.  Since each partition requires at least one database entry (some partitions evaluate to a single game value), the number of partitions provides the tighter lower bound found in Table~\ref{fig:stateSpaceSize}.  This partitioning of both the state-space and set-space also makes our implementation embarrassingly parallelizable, as each partition can be solved independently.

% Setrograde analysis may produce and store as few as one set for each result achieved by any state. In one-suit Bridge, every discrete result $v\in0..d$ is achieved by at least one state. Therefore, with one trick remaining, a minimum of two sets must be stored — one set is needed to describe all deals in which North and South win zero tricks, and one set is needed to describe all deals in which North and South win one trick.  Similarly, for $d=2$ as few as three sets could theoretically describe all deals in $\mathbf{S}_2$.  This bound is quite loose in practice. In Bridge we partition the state-space based on the distribution of suits between the four hands.  This produces the lower bound found in Table~\ref{fig:stateSpaceSize} since each partition requires at least one entry.  This partitioning of both the state-space and set-space is what makes our implementation embarrassingly parallelizable.

\section{Implementation in Bridge}

This section briefly mentions some of the important Bridge implementation details. Further information can be found in the Technical Appendix.

%\JS{Isaac check for correctness. We have some space left, so there is an opportunity to add more material.}\IS{I made small edits, but it looks good other than a small edgecase that just caused a few inconsistent sets to be added to the hypothetical database :) I'm not certain what we should add, but I suspect if we ask someone to read this section, they may ask some useful questions that we can answer.}

\subsubsection{Set Representation:}
% We represent sets using four bits (one each for N, S, E, and W) for each card present in a deal. This representation allows for AND-OR conditions that are not possible with a simple $x$ notation for low cards (e.g., 1100 means North or South could have this card; Figure~\ref{fig-Variable-Placement-set} in the Appendix). \IS{consider moving Fig 3/4 forward, with section 8.1 from appendix}

As discussed in the background section, and depicted in Fig. \ref{fig-x-example}, representing sets of deals with low cards unspecified is not a new idea.  This representation, with fixed cards and \emph{x}'s is used in most (if not all) implementations of Partition Search.  To reduce database size, we extend this methodology, by representing each card using four positional bits — one bit per player — indicating whether or not each player may hold this card.  This representation allows for AND-OR
conditions that are not possible with a simple $x$ notation for low cards.  We demonstrate the utility of this representation by example.
%\IS{AND conditional already exists, OR conditions did too, but we added some.  Feel free to revert to AND-OR as it sounds better an isn't incorrect, or even remove OR and just say "conditions" but I wanted to note this technicality.  \textit{x} representation is North holds some card AND North holds some card AND (North or south or east or west) hold the next card.  What we added is (North or South or west but NOT east).  I would just say "conditions"}

%\JS{Do we really need to use bits to explain the algorithm?} \IS{presumably not, but we don't have another clear concise format to say "this set of players may hold this card, and the remaining players may not." Addressing a later note here as well: \emph{x} refers to a card that may be held by anyone. 1 is that a player may hold a card.  0 is a player is explicitly not allowed to hold a card.  \emph{x} is equivalent to 1111.  using 1 and x would be I think more confusing since it gives x two very different meanings}

\begin{figure}
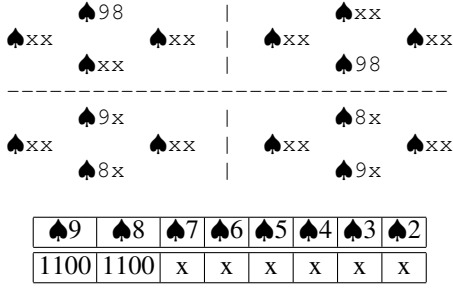

\begin{center}
\begin{tabular}{c}
\begin{lstlisting}
     $\spadesuit$98        |       $\spadesuit$xx 
$\spadesuit$xx        $\spadesuit$xx   |  $\spadesuit$xx        $\spadesuit$xx
     $\spadesuit$xx        |       $\spadesuit$98
-------------------------------
     $\spadesuit$9x        |       $\spadesuit$8x
$\spadesuit$xx        $\spadesuit$xx   |  $\spadesuit$xx        $\spadesuit$xx
     $\spadesuit$8x        |       $\spadesuit$9x
\end{lstlisting}
\end{tabular}
\end{center}

\hspace{2cm}

\begin{center}
\begin{tabular}{|c|c|c|c|c|c|c|c|}
\hline
$\spadesuit$9 &   $\spadesuit$8  &  $\spadesuit$7  &  $\spadesuit$6   & $\spadesuit$5    &$\spadesuit$4  &  $\spadesuit$3  &  $\spadesuit$2\\
\hline\hline
1100&1100&x&x&x&x&x&x\\
\hline
\end{tabular}
\end{center}

\caption{Top: $\spadesuit$98 wins two tricks regardless of the locations of lower cards in each set depicted.
%Players are referenced by cardinal direction, with the North hand depicted at the top, South at bottom, East to the right, and West to the left.
East is next-to-play in all diagrams.
Bottom: Compact representation of the union of the sets above.}
\label{fig-fourSets-example}
\end{figure}

In each of the four sets in Fig. \ref{fig-fourSets-example} (top), North and South can take two tricks by playing the highest two cards, one on each trick.  Since the evaluation of each set is identical (North and South take two tricks on perfect play), the union of the four sets is itself a consistent set.  The union could be expressed as (North holds the $\spadesuit$9 \textit{OR} South holds the $\spadesuit$9) \textit{AND} (North holds the $\spadesuit$8 \textit{OR} South holds the $\spadesuit$8) \textit{AND} (all lower cards are \emph{x}'s).  Using 4 positional bits per-card (in North-South-East-West order) we can represent that statement compactly.  The $\spadesuit$9 and $\spadesuit$8 each have their bits set to true (1) corresponding to North and South holding those cards, and two bits false (0) corresponding to East and West set not holding those cards, as illustrated in Fig. \ref{fig-fourSets-example} (bottom).  Our databases consist of entries mapping from sets represented in this syntax to a bound on the number of tricks taken by each partnership.

% A bitwise ``or'' of two Sets with the same value will produce a more general set.
%\IS{A bitwise or of two sets that are identical apart from a single card will produce a more general \emph{and} consistent set.  A bitwise ``or'' of sets that diverge by more than one card will produce a very general set, but even if they have the same value, this set may contain states that weren't in either original set, and therefore may not be consistent.  Obviously I did some work here to do better than just merging sets that are off by one card, but it's actually not obvious that it's particularly important to do that in Bridge.  Anyway, there's a lot of complexity here and I think there's still a lot of work to be done around set representation and merging.  I experimented with generalization that duplicates information as much as possible without increasing the tree size (sometimes allowing pairwise merges that otherwise would not occur), and group-wise merging instead of pairwise (all of which I got to work with varying degrees of computational cost and storage benefit) but I am currently running one of the simplest versions, and making up for it by performing negative queries (not x is not in database) of the database instead of positive queries (x is in database).}

\subsubsection{Low-Cost Generalization:}
State generalization uses binary search. The lowest card in a suit can always be marked as $x$ and it is possible that entire suit (13 cards) could be $x$s. Hence for each suit a binary search is done on the number of $x$s. The program starts in the middle of the range and introduces that number of $x$s. Depending on whether the resulting set is consistent, the search either tries adding more or eliminating some $x$s.  In some cases, the 4-bit representation used can compactly express the union of sets in the database, reducing database size.  Compaction operations can be performed at insertion or in post-processing (Alg. \ref{alg:Setrograde2}, line 12). %Once the database is mostly generated, we use a compactEDB function to combine set taking advantage of the more expressive representation we use for storage.\IS{todo: reword}% \NS{Can we say something about generating AND-OR conditions?}\IS{I'm not quite sure what you mean here.  I'm guessing you're asking about the set compaction, which expresses multiple sets as a single set.  We could talk about that here, though that is a separate step.  It's perhaps more related to set-representation directly.  AND conditions exist between each card in the 4-bit set representation. OR conditions are internal to a single card (each card may belong to A or B or C, AND each other card has its own set of (or) constraints.}

\subsubsection{Low-Cost Querying:}
State lookups, matching a state to a set to retrieve a value, are achieved using a depth-limited tree data structure. At each node, a bitwise AND operation can be used to determine whether a state is in a set.  The tree structure lets us minimize the nodes that need to be tested.  Each node contains partial information, and if a node does not provide a partial match, its subtree can be pruned.  The tree we use benefits from locality, and we maintain relatively small independent trees for each distribution of suits between players (Lower Bound in Table \ref{fig:stateSpaceSize}).
% States can be matched against a set with bit operations (a benefit of the set representation discussed above). A successful partial match results in following a branch; a failed partial match looks for matches at other branches of the tree.\IS{continues to iterate through the tree; low depth; high branching; average nodes in lookup?} %\JS{To discuss. I thought that the other branches were part of the tree node. }\IS{functionally, kind of.  In practice, it is more like you know where the first child is, and the children are a linked list, and each child may have children of it's own (another linked list) each of which may have more children.  We just use implicit pointers for children, and explicit pointers for siblings.}
%at additional information at that tree node.%\IS{this is fine, but on a technicality, a failed match does not go down another branch, it goes sideways.  It is possible the search will have to retreat upward and continue the search in another branch.  We use a very shallow and wide tree.  Incidentally, future work may wish to include using variable card orderings on each branch of the tree}

\subsubsection{Disk Storage:}
The 24-card database is partitioned into $10^6$ independent pieces (Lower Bound in Table~\ref{fig:stateSpaceSize}), each one reflecting a different distribution of cards (the deal's $shape$). Within each partition the sets are organized in a tree-like fashion. More details are in the Technical Appendix.

\begin{algorithm}[t]
\caption{Setrograde Helper Functions}
\label{alg:setrogradeHelpers}

\begin{algorithmic}[]
    \STATE \# Returns value of a state.
    \STATE \textbf{function} \textit{databaseLookup}(state $s$, distance $d$)
    \begin{ALC@g}
        
        \RETURN $\max_{s' \in \text{succ}(s)}EDB_{d-1}[s']$
    \end{ALC@g}
    \STATE \textbf{end}
    \STATE

   \STATE \# Returns a state that is not yet present in the EDB.
    \STATE \textbf{function} \textit{nextIndependentState}(set of set of states $\mathbf{T_d}$)
    \begin{ALC@g}
            \IF{$\exists s\in\mathbf{S}_d|s\notin \mathbf{T_d}$}
                    \RETURN $s$
            \ENDIF
            \RETURN \textbf{null}
    \end{ALC@g}
    \STATE \textbf{end}
    \STATE

    % \STATE \textbf{function} databaseLookup(state $s$, distance $d$)
    % \begin{ALC@g}
    %     \STATE $res \gets worstOutcomeForPlayer()$
        
    %         \JS{function has side-effect of maxing over res} \IS{yes, it's a one-step minimax.  Should I try to write it in math as a one-liner?}
    %         \FOR{\textbf{all} next state s'}
    %             \STATE $res \gets best(EDB_{d-1}[s'],res)$

    %             \JS{$d$ should be a paramter} \IS{resolved in declaration and call}
    %         \ENDFOR
    %         \RETURN $res$
    % \end{ALC@g}
    % \STATE \textbf{end}
    \STATE \# Returns whether all states in a set share given value.
    \STATE \textbf{function} ORACLE(set of states t, value v)
    \begin{ALC@g}
        \RETURN $ \not\exists s \in t | databaseLookup(s) \neq v$
    \end{ALC@g}
    \STATE \textbf{end}
    
    \STATE
    \STATE \# Returns a collection of candidate sets
    \STATE \textbf{function} generateCandidates(state s)
    \begin{ALC@g}
            % \RETURN $ \{t\in \mathbf{T} | s\in t \}$
            \RETURN $ \{t\in 2^{\mathbf{S}_d} | s\in t \}$ 
    \end{ALC@g}
    \STATE \textbf{end}
    
    \STATE
    \STATE \# Returns a consistent set.
    \STATE \textbf{function} generalizeToSet(state s, value v)
    \begin{ALC@g}
            \STATE $t_{ret} \gets s$
            \STATE $candidates \gets generateCandidates(s)$
            \WHILE {$ \exists t_{c} \in candidatates\;s.t. |t_c| > |t_{ret}|$}
                    \IF{$ORACLE(t_c, v)$}
                        \STATE $t_{ret} \gets t_c$
                    \ENDIF
            \ENDWHILE
            \RETURN $t_{ret}$
    \end{ALC@g}
    \STATE \textbf{end}
    
\end{algorithmic}

\end{algorithm}

\section{Experimental Results}
Here we provide an analysis of setrograde analysis performance on Bridge database generation.
The program is written in Julia, and compiled in version 1.8 or higher using the LLVM compiler. All code is compatible with version 1.11.  %Compiler is LLVM (Julia compiler.  typically compiled with O3).  Code is just-in-time compiled.
% \subsection{Hardware and Software Details:}
The large databases were computed on a machine with 48 cores, 187 GB of RAM, and 256 GB of swap using an Intel(R) Xeon(R) Gold 6248R CPU @ 3.00GHz.

%2: M1 chip (2021 macbook pro)
%3: Eureka; 140 ish cores? not much RAM or SWAP; used for initial run of BFOG size 5.
%M1 solved size 3 in 7.5 seconds, draco solves size 3 in 10.7 seconds. (both single threaded
%Draco solves size 5 in 1 hour (~50 cpu hours)
%Eureka solved size 5 in 2 cpu years (about 4-5 days); older form of the algorithm

% Size 6: 6 days on DRACO (5 if we run the fastest formulation, but then we get 56GiB instead of 50GiB so I'm giving the numbers for 50GiB since one day isn't a big deal).

\subsection{Validation}

Extensive tests to verify the correctness and completeness of our databases were performed. The simplest approach is to compute the retrograde analysis databases and then compare the results with the setrograde data.
%to build a database using retrograde analysis with a standard solver.  Our implementation of a state-wise retrograde search uses a minimax solver and acts as a baseline for correctness. The minimax solver was validated on 2000 Double Dummy problems against the widely used Haglund double dummy solver solver.  The standard solver is slow on full-sized deals, limiting the scope of this experiment.
%
%We also wrote a Double Dummy solver that implements Partition Search.  The Partition solver was validated against the Haglund solver, and additional analysis was performed at each node on a 2000 problem test set using our minimax solver to verify correctness throughout the tree.
%To the extent that computational resources allowed, we generated databases using standard retrograde techniques.
Retrograde databases were built through 16-card deals. Every state's value in these databases was in agreement with the corresponding setrograde result.
%While it was prohibitively expensive to generate and store the solution to all 16-card deals using standard methods, we generated the data in place over 36.5 hours on a Macbook Pro M1.  At each step, a lookup was performed in a setrograde database to verify the correctness and completeness of the setrograde database.  This experiment was performed twice.  All further databases were validated against the verified setrograde databases.  Cross-validation takes just 2 hours for 16-card deals.

%\JS{Isaac, I simplified the discussion. Is the Second validation essentially correct?}\IS{yes, it is.  There are a few other small things we could mention, but I think this is a reasonably strong validation argument, and the most important validation methods are present.}
Beyond 16 cards, exhaustive validation is no longer practical.  We rely instead on two methods for partially validating the 20 and 24 databases. First, random subsets of the search space were chosen and for each state the setrograde value was compared to that of a standard minimax search. Second, thousands of double dummy problems were solved using a search-based solver with and without using the setrograde databases. The values returned in all the searches were identical.

Our results were validated to the extent that could reasonably be done on the available compute resources. Positive results have been achieved using every validation method we could practically implement.

\subsection{Performance of Setrograde Analysis}

Table~\ref{fig:stateSpaceSize} shows one measure of setrograde's performance. Upper Bound is the number of states that a retrograde analysis program would considered (in the reduced search space). Setrograde is the number of sets that were needed to capture the exact same information. There is a $4 \times 10^4$ reduction. This is slightly misleading as the cost to compute the value of a state is much less than the cost of producing a consistent set.

In Table \ref{fig:memStore} the storage and computational resources for retrograde and setrograde analysis are presented.
Some of the retrograde numbers were too costly to run and their values are extrapolated (indicated by a \dag). The retrograde storage is pessimistic (two states per byte), given the potential for applying further data compression techniques. The computation times for retrograde are optimistic as they do not take into account performance loss due to scaling (e.g., loss of locality).

\begin{table}
\small
    \centering
    \begin{tabular}{crrrrrrrrrrr}
    \toprule
& \multicolumn{2}{c}{DB Size (GiB)} &
\multicolumn{2}{c}{Gen Time (CPU Days)}&\multicolumn{1}{c}{States/Byte}\\
% \multicolumn{4}{c}{Setrograde}
\cmidrule(l){2-3} \cmidrule(l){4-5}\cmidrule(l){6-6}
 Cards & \multicolumn{1}{c}{Retrograde}& \multicolumn{1}{c}{Setrograde}& \multicolumn{1}{c}{Retrograde}& \multicolumn{1}{c}{Setrograde}& \multicolumn{1}{c}{Setrograde}\\
  % & \multicolumn{1}{c}{DB Size (GiB)} &\multicolumn{1}{c}{DB Size (GiB)} & \multicolumn{1}{c}{CPU Days} & \multicolumn{1}{c}{CPU Days}\\
 %size  & deals & \multicolumn{1}{c}{deals} & entries & bytes & per byte & CPU time & bytes & per byte & CPU time \\
% \midrule
% \multicolumn{7}{c}{Retrograde}\\
\midrule
         4&   $1\times10^{-7}$ & $7\times10^{-6}$&       —&—&$2.6\times10^{-2}$\\
         8&   $4\times10^{-5}$ & $3\times10^{-5}$&       —&—&$3.2\times10^{0}$\\
         12&  $2\times10^{-2}$ & $1\times10^{-3}$&       $3\times10^{-3}$&$1\times10^{-4}$&$3.1\times10^{1}$\\
         16&  \dag $5\times10^{0}$ & $5\times10^{-2}$& $1\times10^{0}$&$1\times10^{-2}$& $1.9\times10^{2}$\\
         20&  \dag $2\times10^{3}$ & $2\times10^{0}$& \dag$4\times10^{2}$ & $2\times10^{0}$ & $1.6\times10^{3}$\\
         24&  \dag $4\times10^{5}$ & $5\times10^{1}$& \dag$1\times10^{5}$&$3\times10^{2}$&$1.5\times10^{4}$\\
\bottomrule
% \midrule
% \multicolumn{7}{c}{Setrograde}\\
% \midrule
%          1&   $3\times10^{7}$&$2\times10^{2}$& $7\times10^{1}$& 7K&0 &0\\
%          2&   $9\times10^{12}$&$9\times10^{4}$& $8\times10^{3}$& 27K&3 &0\\
%          3&   $4\times10^{17}$&$3\times10^{7}$& $5\times10^{5}$& 1.1M&26 &10(s)\\
%          4&   $3\times10^{21}$&$1\times10^{10}$& $3\times10^{7}$& 51M&187 &16(m)\\
%          5&   $7\times10^{24}$&$3\times10^{12}$& $1\times10^{9}$& 1.7G&1644 &50(h)\\
%          6&   $5\times10^{27}$&$8\times10^{14}$& $2\times10^{10}$& 50G&14901 &288(d)\\
%          % 7&  4e16& 1e30&2e17& ?& ?&?
%           % &est: 55 (y)
%            \bottomrule
 \end{tabular}
\caption{Database Generation Time and Storage for Bridge}
\label{fig:memStore}
\end{table}

\textbf{Database Size:} The retrograde database sizes reported used $0.5$ bytes per state, indicating a deal's value $0..d$.
This representation reflects having a function that maps a state to a unique storage location with no gaps. Although we did not use such a function (it was too slow), the results are reported as if we did.
%\IS{this is fine, but *technically* not done.  The mapping function exists, but is too slow to be practical, and we haven't stored the 4-bit version.  This seems innocuous, but is a slight inaccuracy regarding work done.  The 4-bit version is technically a hypothetical, albeit one with a clear implementation}
Additional compression techniques (not done) could yield small gains. Retrograde database sizes scale linearly with the number of states in the state-space.

In contrast, setrograde has to store variable size sets (for 24 cards, ranging from 8 to 40 bytes, but compression gives a factor of 5 reduction).
%\IS{8*(tricks remaining -1) BUT with a decent amount of empty space, so compression is incredibly effective (factor of 5 about) and a deep entry will usually have siblings relying on the same parents as it, decreasing the effective size of each entry}).
Despite the additional storage for a set, our setrograde database is almost 4 OOM smaller than its retrograde counterpart (before investigating data compression techniques for the retrograde counterpart).%\IS{Clever solutions probably exist to reduce set storage size by at least half}
%\JS{Just checking... Table 1 says 4.4 OOM reduction in states to sets. Storage says roughly a 4 OOM. Does that mean that states are only 4 times bigger than states in the DB?}\IS{you mean sets in the DB, and yes, I can triple check, but the database contains on average less than 2 64-bit integers per set stored prior to compression, and compression gets it down pretty far. 
 % The set representation is not particularly large, especially when considering the tree structure where information is scattered along the path}

%Retrograde storage could be reduced by a constant factor using known techniques such as implicit state representation using an analytical ranking function, however, this optimization is difficult in Bridge and a constant factor (even a factor of 12.5 — which can be achieved by representing the value of each state using 4 bits) has limited utility in allowing us to scale to larger depths. \JS{I would change the above. With an implicit scheme you would use 4 bits per state. Use that for computing the size of the DB. State a caveat that there could be additional techniques to get more compression.}  These techniques do not apply to setrograde analysis, where sets must be stored explicitly.  Entries marked with \dag are estimates based on linear scaling with respect to the state space, and are included for large state spaces that were not practical (or possible) to solve on available compute resources.

\textbf{Generation Time:} The time taken to generate a database is presented in number of CPU days. Note that the machine used has 48 cores, meaning one day of wall-clock time corresponds to 48 CPU days.

Through 12 cards, the execution time for both retrograde and setrograde analysis are too small to draw meaningful conclusions. For the 16-card calculation, setrograde is completed with 2 OOM less time. The 20 and 24 card retrograde computations were not performed. Both retrograde and setrograde analysis are embarrassingly parallelizable for Bridge. The setrograde wall clock times for 20 and 24 cards were 1 hour, and 6 days respectively (on a 48 core machine).%\JS{Why was the 20 retrograde not performed? The numbers above say it would be done in 6 days.}\IS{because we do not have a stored 16 card retrograde database, and that's the theoretical number assuming memory is a complete non-issue and non-factor in runtime. We would be performing queries in a 50GB hash table at that point, which I could break up, like I had to do for setrograde at the larger sizes. but even then it's a massive task, and the database we produce would not get saved, or if it was would take a major engineering effort to get working.  The simplest and most accurate reason is that I could either do this, or I could do everything else I did.  There was a lot of code, and there were a lot of bottlenecks and engineering challenges.  I prioritized the ones that got us further.  Keep in mind, the actual database sizes for retrograde are significantly larger, because there's again a large engineering effort involved in constructing a piecewise ranking function that removes symmetries (or we add back symmetries and get the 4 bits, but lose a factor of 24, so we end up right about where we started size wise)}
%\JS{Computation time for 24 is less than 3 OOM, assuming one believes your extrapolated time.}\IS{the extrapolation is probably optimistic}

\textbf{States/Byte:} The number of states divided by the size of the resulting database is a measure of information density. A  retrograde solver would store roughly $2$ states per byte (somewhat higher with appropriate compression), regardless of the value of $d$. For setrograde the information density grows with $d$. Increasing $d$ means that a set, in general, reflects a larger number of states.

\subsection{Breakdown of 24-card performance gains}
Setrograde analysis decreases both storage and computation costs, rather than trading one for the other.  To understand where the computation and storage costs are being reduced, we can break down the performance of setrograde analysis.  In the 24-card case there are $8\times 10^{14}$ effective states. This is reduced to $2\times 10^{12}$ states added to the open list (generated) throughout the database generation process — a reduction of 2.4 OOM. The open list never exceeds 1000 states.
% \JS{But you never have $2.4 \times 10^{12}$ states on the OL at any time. What was the maximum size of the OL when computing the 24-card DB? Does this mean that the algorithm really is running at O($10^{12})$, a 2 OOM improvement?}\IS{I don't have the max size of open list for size 6, but for size 5, it doesn't seem to exceed 100 states.  Basically the open list is maintained more or less as a depth-first record, so the closed list and even the database are guaranteed to grow much faster than the open list.  Syntactical clarification: throughout the entirety of the search, $2\times10^{12}$ states are added to open; 46 out of 47 are duplicates; and at any given time, the open list is quite small.  We don't know how small for size 6, but I stopped tracking it because it wasn't a bottleneck.  ALSO, the algorithm is running better than a 2 OOM improvement, because duplicate detection catches 46 out of 47 states on open, and is much faster than evaluation and generalization.  Edited the text slightly for clarity} 
Of the $2\times 10^{12}$ states placed on the open list, only 1 in 47 is independent. The remaining 46 states are discarded using (cheap) duplicate detection techniques ($\approx5\%$ of total runtime) leaving us with just $4.6\times10^{10}$ states to evaluate and generalize.

Over $90\%$ of computation time is spent evaluating and generalizing the $4.6\times10^{10}$ independent states.  Each evaluation and generalization step results in adding a set to the final database. A post-processing phase (\emph{compactEDB} in Alg.~\ref{alg:Setrograde2}) scans the database to identify sets that can be combined to form a single, more general set.
%The database undergoes a final round of compression in which set representations are compacted \JS{How? Are some sets subsumed by later sets?}\IS{combined, not subsumed.  I find pairs or groups of sets that can be re-written as a single set.},
This is a small additional computation cost that results in roughly cutting the final database in half to $2.4\times10^{10}$ sets.
%\JS{I do not see  this adding value. Start}The final round of compression impacts storage, but has no positive impact on computation (unless you consider I.O., which very well may take more resources than set compaction).  However, the reduction from the initial statespace down to the number of sets evaluated is more relevant to computation than it is to storage, since even after the number of evaluate and generalize steps is reduced by a factor of $2\times10^{4}$, those steps still dominate the computation. \JS{End}\IS{I think it's useful to clarify that the final round of compression impacts storage but not computation, but the rest of is I'm not sure is neccesary.  The reason I include that is to emphasize where the storage gains and where the computation gains are coming from, but we can remove it.}

Setrograde's generalize step has no counterpart in retrograde analysis -- and it is expensive. On average, it increases the cost of evaluating a state by a factor of 15 in our implementation. This is a high price to pay but, of course, it leads to a 4 OOM reduction in the number of states considered for 24 cards. Our generalize function might still be made less expensive; we have not definitively determined the best way to maximize set generality and minimize computation costs.

\subsection{Impact on Double-Dummy Search}

% \ORG{Isaac to generate these over the weekend.}
Our long-term aim is to build a 52-card database that can be used directly in Bridge playing algorithms.
% Although it is not our primary intention to use the endgame databases as part of a double-dummy search, we envision a 52-card database that can be used directly in Bridge playing algorithms.
In the short term, we have performed experiments with a DD solver which show that the 6-trick database can eliminate roughly 75\% of the tree, and the 7-trick database (when available) will push this up to roughly 90\%. %Note that state-of-the-art solvers rely on considerable expert knowledge to make the search manageable. Setrograde requires no expert knowledge; it is just a database lookup.\IS{Last two sentences feel awkward.  I know why we have this section, but it still feels off-balance to me.  The goal is 1) reach 52 cards, at which point I get to work on the auction (yay) and 2) apply this algorithm to other applicable domains where it can make domains that were previously considered unapproachably large approachable (or scale domains, i.e. from 6x6 to 7x7 or 8x8 chinese checkers, larger rubiks cubes or STPs, etc...).  I don't think we should omit this entirely, I just want to find a way to re-write it where it doesn't feel like we are trying to justify that our fish climbs trees, even if it has potential to be a pretty decent tree-climbing fish.}

\section{Conclusions}

This paper introduces setrograde analysis, a generalization of retrograde analysis from states to sets. For applicable domains, the algorithm can reduce the computational and storage needs by orders of magnitude. Some games for which setrograde analysis will be beneficial include Chinese Checkers and Skat.

For the game of Bridge, endgame databases have not been built because the massive search space and storage needs made it seemingly impractical. Setrograde makes this possible through 24 cards. The 28-card databases are currently being computed, and 32 card databases should be possible with today's technology. The growth rate of the resource needs seems to be rapidly decreasing with size, making it now possible to imagine solving the entire 52-card deal space.

\section{Acknowledgements}
This work was funded by the Canada CIFAR AI
Chairs Program. We acknowledge the support of the Natural Sciences and Engineering Research Council of Canada
(NSERC).

\bibliography{references}

\newpage
~\newpage

\section{Technical Appendix}

The technical appendix goes into the implementation details of our setrograde analysis Bridge program.
For clarity, our examples are limited to using one suit (spades). The generalization and iteration processes described in Alg. \ref{alg:setrogradeHelpers} are augmented with examples. A description of the tree structure used to store our databases is described.

\subsection{Generalizing and Iterating in Bridge}

%\IS{Key points: 
%- we describe low cost iteration in section 3 (example 2), and that is the most in-depth we get
%- we need to say, multiple ways, but here's the approach we used when generating the 6-card databases (canonical iteration with a high frequency of contiguity of states; it's really a canonical ordering with branches where branches are not independent, and duplicate detection within the next-independent state function prunes the duplicates)
%- we describe low-cost generation quite well in section 5.  In fact, the reference from section 3 while correct, could probably have pointed toward section 5 rather than the appendix.  The key point here is to reference section 5 from the appendix (along with section 3.2, which is where we point toward the appendix) and quickly segue into the illustrative examples
% }
% \IS{from here down I need to do a quick pass to revise terminology to match everything above; This was never revised after the last meeting prior to Uganda}
Alg.~\ref{alg:Setrograde2} makes use of two functions that require careful implementation to scale well ($nextIndependentState$ and $generalizeToSet$). A high-level version of these routines can be found in Alg.~\ref{alg:setrogradeHelpers}.
In this section we describe one implementation for each that performed well on 24-card deals.

In Alg. \ref{alg:Setrograde2} we iterate sequentially through $d\gets1..D$. For each retrograde distance, we add states to the open list, evaluate them, generalize them (if not a duplicate), and remove them from the open list. For each non-duplicate state, the generalization process will result in 0 to 4 states being added to the open list. By greedily pulling states nearest the end of the canonical ordering from the open list, we can minimize the expansion of the open list.  The open list's memory footprint is dominated by the disjoint segments of the database.
% \IS{Not sure that it's easier to follow, but I think it's more to the point} States are pulled from the open list in a depth-first manner — that is, the state that occurs latest in the canonical ordering is evaluated first. Any states added to the open list following this evaluation will be later in the canonical ordering than the state pulled from open.  This prioritizes generating and evaluating states whose generalization will result in no open-list additions.\IS{this is NOT the same as narrowness, since we are happy to generate 4 successors instead of 1, as long as we get closer to a leaf state (state with no successors); I'm having some trouble expressing that clearly}
% The open list size expands and contracts in a manner similar to depth fluctuations in a depth-first search, and remains small enough that it should not become a performance consideration.  
In the following, the iteration process is described in more detail.
 
 % by We perform a binary search on the size of $\mathbf{S}_{input}|d_1\in \mathbf{S}_{input}$, querying the oracle until we have found the set for which the oracle returns \textbf{True} or \textbf{False} with the fewest specified cards, $\mathbf{S}_{best}$.  This set is mapped to a lower bound or an upper bound on the number of tricks North and South take, and inserted into a solution tree, which we maintain as a specialized data structure.  Rather than repeating this for each deal, we skip from $d_1$ to the next independent deal $d_k|d_k \notin \mathbf{S}_{input}$.  If no such deals exist, each deal in $\mathbf{S}_d$ must be lower bounded by $R$ or upper bounded by $R-1$ in the solution tree, so we increment $R$.

 To enable fast iteration we define a canonical ordering of suit permutations, $s_1 ... s_n$ such that any set that may be represented using fixed ranks and \emph{x}'s will consist of contiguous states in that ordering. To do that we define a process for trading cards between hands where low cards permute before higher cards.  To find the next state in the canonical ordering from a set of states, the high cards are incremented by one permutation, and the \emph{x}'s are reverted to the first permutation.

In the generalization process, the lowest card in a suit can always be marked as \emph{x}. 
 Therefore, anywhere from one to all of the cards in a suit could be \emph{x}'s.  With at most 13 cards in a suit, there are at most 13 configurations of \emph{x}'s in a suit on any deal.  This is reflected in the binary search we described in Section 5 where initially one more than half of the cards in a state are marked as \emph{x}.  For each set considered, a query is made to the oracle (Alg.~\ref{alg:setrogradeHelpers}) to check whether the set is consistent.  If a set is consistent, we place a lower bound on the number of \emph{x}'s we will have in our ``best'' set.  When a set is inconsistent, we place an upper bound instead.  This process is repeated until a tight bound is obtained, and the consistent set with the most \emph{x}'s is added to the database.
 
 Here, we work through several examples.  With 4 cards remaining, the canonically first deal is generated.  The deal is evaluated (North and South take 1 trick), and \textit{generalizeToSet} is called. In the following examples, a \textbf{T} (for \textbf{True}) indicates that a set is consistent, and \textbf{F} (\textbf{False}) that it is not.

\begin{center}
\begin{tabular}{c}
\begin{lstlisting}
   $\spadesuit$5          $\spadesuit$5           $\spadesuit$x
$\spadesuit$2     $\spadesuit$3 : $\spadesuit$x  $\mathbf{T}$   $\spadesuit$x -> $\spadesuit$x $\mathbf{F}$ $\spadesuit$x
   $\spadesuit$4          $\spadesuit$x           $\spadesuit$x
\end{lstlisting}
\end{tabular}
\end{center}

We start by marking three cards as \emph{x}. The lowest card is always an \emph{x}, so the minimum is 1 and the maximum is 4. The midpoint --2.5-- rounds up to 3 (performance is marginally better when rounding up). When the initial oracle query returns \textbf{True}, we know the set is consistent, and that we will have at least 3 \emph{x}'s. The oracle is queried again on a larger set (4 \emph{x}'s). On the larger set, the oracle returns \textbf{False}.  We now have a tight bound on the number of \emph{x}'s in the ``best'' consistent set.  The ``best'' consistent set ($\mathbf{t}_{ret}$ in Alg.~\ref{alg:setrogradeHelpers}) is the deal with $\spadesuit5$ specified and all other cards \emph{x}'s.  It is added to the database, and the next deal is generated.  \textit{nextIndependentState} permutes the high cards (just the $\spadesuit5$ in this case) and sets the low cards back to their first permutation.  To complete the 4-card database, we repeat the process on the following 3 deals:

\begin{center}
\begin{tabular}{c}
\begin{lstlisting}
   $\spadesuit$4          $\spadesuit$x           $\spadesuit$x
$\spadesuit$2     $\spadesuit$3 : $\spadesuit$x  $\mathbf{T}$   $\spadesuit$x -> $\spadesuit$x $\mathbf{F}$ $\spadesuit$x
   $\spadesuit$5          $\spadesuit$5           $\spadesuit$x
\end{lstlisting}
\end{tabular}
\end{center}

\begin{center}
\begin{tabular}{c}
\begin{lstlisting}
   $\spadesuit$4          $\spadesuit$x           $\spadesuit$x
$\spadesuit$2     $\spadesuit$5 : $\spadesuit$x  $\mathbf{T}$   $\spadesuit$5 -> $\spadesuit$x $\mathbf{F}$ $\spadesuit$x
   $\spadesuit$3          $\spadesuit$x           $\spadesuit$x
\end{lstlisting}
\end{tabular}
\end{center}

\begin{center}
\begin{tabular}{c}
\begin{lstlisting}
   $\spadesuit$4          $\spadesuit$x           $\spadesuit$x
$\spadesuit$5     $\spadesuit$2 : $\spadesuit$5  $\mathbf{T}$   $\spadesuit$x -> $\spadesuit$x $\mathbf{F}$ $\spadesuit$x
   $\spadesuit$3          $\spadesuit$x           $\spadesuit$x
\end{lstlisting}
\end{tabular}
\end{center}

\noindent
resulting in the three consistent sets (middle) being added to the database.

Though we perform our computation on the fixed-rank representation, entries are stored in the more expressive 4-bit format described in the paper. During the solution insertion process, or optionally in post-processing, sets are further compacted, leveraging positional bits to combine sets. The complete 4-card database contains the following 2 entries, representing the 24 deals with 4 cards remaining in a single suit:

\begin{center}
\begin{tabular}{|c|c|c|c|c|}
\hline
Tricks& $\spadesuit$5    &$\spadesuit$4  &  $\spadesuit$3  &  $\spadesuit$2\\
\hline\hline
1 &1100&x&x&x\\
\hline
0 &0011&x&x&x\\
\hline
\end{tabular}
\end{center}

With a complete database for 4-card deals, it is possible to generate the solutions to 8-card deals similarly.  There are 2,520 deals with 8 spades remaining, and East to play. These are compressed into 19 database entries.
%\IS{omitting longer tables here — if we think they add a lot, we can, but realistically, even the most dedicated reader will not parse out the bits beyond the 1-trick table.}%In the table below $\spadesuit2,3,4$ are omitted since they are always \emph{x}'s:

% \IS{Note: placeholder table.  Final table would be 19 lines. It's printed in a very non-readable format (ordering of bits is difficult to read, but convenient for code; player order as well.  Current table is incorrect)}

% \JS{Explain LB and UB. Why use 9/8/7/6 instead of the normalized 5/4/3/2}\IS{If we include a table of this form, I will use tight bounds instead to make it easier for the reader. It is normalized, but 5/4/3/2 are all x's;}

% \IS{Table commented here}
% \begin{center}
% \begin{tabular}{|c|c|c|c|c|c|c|}
% \hline
% LB & UB & $\spadesuit$9 & $\spadesuit$8    &$\spadesuit$7  &  $\spadesuit$6  &  $\spadesuit$5\\
% \hline\hline
% % 2 &  &1100&1100&x&x&x\\
% % \hline
% 2 &  &1100&1100&x&x&x\\
% \hline
% 2 &  &0101&0100&0001&x&x\\
% \hline
% 2 &  &0001&0100&0011&0011&x\\
% \hline

%  & 1 &0011&x&x&x&x\\
% \hline
%  & 1 &1011&0011&x&x&x\\
% \hline
%  & 1 &0001&1011&x&x&x\\
% \hline
%  & 1 &0001&1100&1100&x&x\\
% \hline
%  & 1 &0001&1100&1110&1100&x\\
% \hline

%  & 0 &0011&0011&x&x&x\\
% \hline
%  & 0 &0011&0111&0011&x&x\\
% \hline
%  & 0 &1000&0001&1000&x&x\\
% \hline
%  & 0 &1000&0001&1000&x&x\\
% \hline

% \end{tabular}
% \end{center}

We provide a diagram of the generalization process for the cannonical first 8-card state, similar to the 4-card example above.  We also show the next independent state ($sNext$). A binary search is done on the number of $x$'s (start with 5 (\textbf{True}), jump to 7 (\textbf{False}), and then back down to 6 (\textbf{True}). We now have the maximal consistent state (by number of $x$'s). It is then permuted (moving the 8 from the N hand to the S hand in this case), and that set is then selected from that set.

\begin{center}
\begin{tabular}{c}
\begin{lstlisting}
   $\spadesuit$98            $\spadesuit$98            $\spadesuit$9x
$\spadesuit$32     $\spadesuit$54 : $\spadesuit$xx  $\mathbf{T}$   $\spadesuit$xx -> $\spadesuit$xx $\mathbf{F}$ $\spadesuit$xx
   $\spadesuit$76            $\spadesuit$7x            $\spadesuit$xx
\end{lstlisting}
\end{tabular}
\end{center}

\begin{center}
\begin{tabular}{c}
\begin{lstlisting}
      $\spadesuit$98            $\spadesuit$9x             $\spadesuit$97
-> $\spadesuit$xx  $\mathbf{T}$  $\spadesuit$xx : $\spadesuit$xx        $\spadesuit$xx : $\spadesuit$32 sNext $\spadesuit$54
      $\spadesuit$xx            $\spadesuit$8x             $\spadesuit$86
\end{lstlisting}
\end{tabular}
\end{center}

%\IS{ we can consider adding the number of deals in this sample set and a small discussion instead of just the numbers for the complete database}

Once the 8-card database is complete, the process is repeated to generate the 12-card database.  The 12-card database consists of 295 entries, representing 369,600 deals.

% \JS{Need to talk about the size of the open list...
% In the 24 card computation, the open list does not reach 1,000 entries.}

With the introduction of the other three suits, the approaches described above for iteration and generalization need only minor adjustments.
\begin{itemize}
    \item 
    In generalization, suits are \emph{not} independent, therefore we must perform binary search on each suit recursively (i.e. we could perform $log(|\clubsuit|)\times log(|\diamondsuit|)\times log(|\heartsuit|)\times log(|\spadesuit|)$, rather than a sum of the same terms).
    \item
    In the iteration process, we may add up to 4 elements to the open list at each step to ensure we cover the entire state-space.  Each successor is constructed by applying the permutation process we described to a single suit, while the other suits remain untouched. This may cause some duplication.
    \item
    Queries and storage are largely unaffected by the addition of suits; rather than representing cards in a high to low order, an ordering of all the cards (prioritizing high cards regardless of suit) is used.
\end{itemize}

\subsection{Storing Setrograde Databases}

This section describes how the compact sets described in Sec. 5 of the paper are stored to support fast-enough lookups with low-enough memory overhead. 

This representation is likely \emph{not} the most compact format possible, nor provides the fastest possible lookups.  At the time of this writing, the program is computationally limited; our largest databases fit on consumer-grade thumb drives. Therefore, our attention has been on decreasing computation costs. Database queries, while expensive due to frequency, are not (yet) dominating our computation, which is how we choose to define ``fast-enough''.  The implementation described was chosen for its simplicity and empirical success in supporting sufficiently fast queries and relatively small disk footprints.

% It is important to discuss your data structure, beyond what is in the main body of the paper. 
% - chose your implementation for simplicity
% - may be more compact ways to represent it
% - allows easy manipulation of AND and OR relationships

% The databases are not as compact as they could be.  Storage is not the bottleneck, so reducing them in size by, say, a factor of 2 is somewhat irrelevant.  As we move to solving a larger number of cards, we may have to revisit the database design.

\begin{figure}
    \centering
    \includegraphics[width=1.0\linewidth]{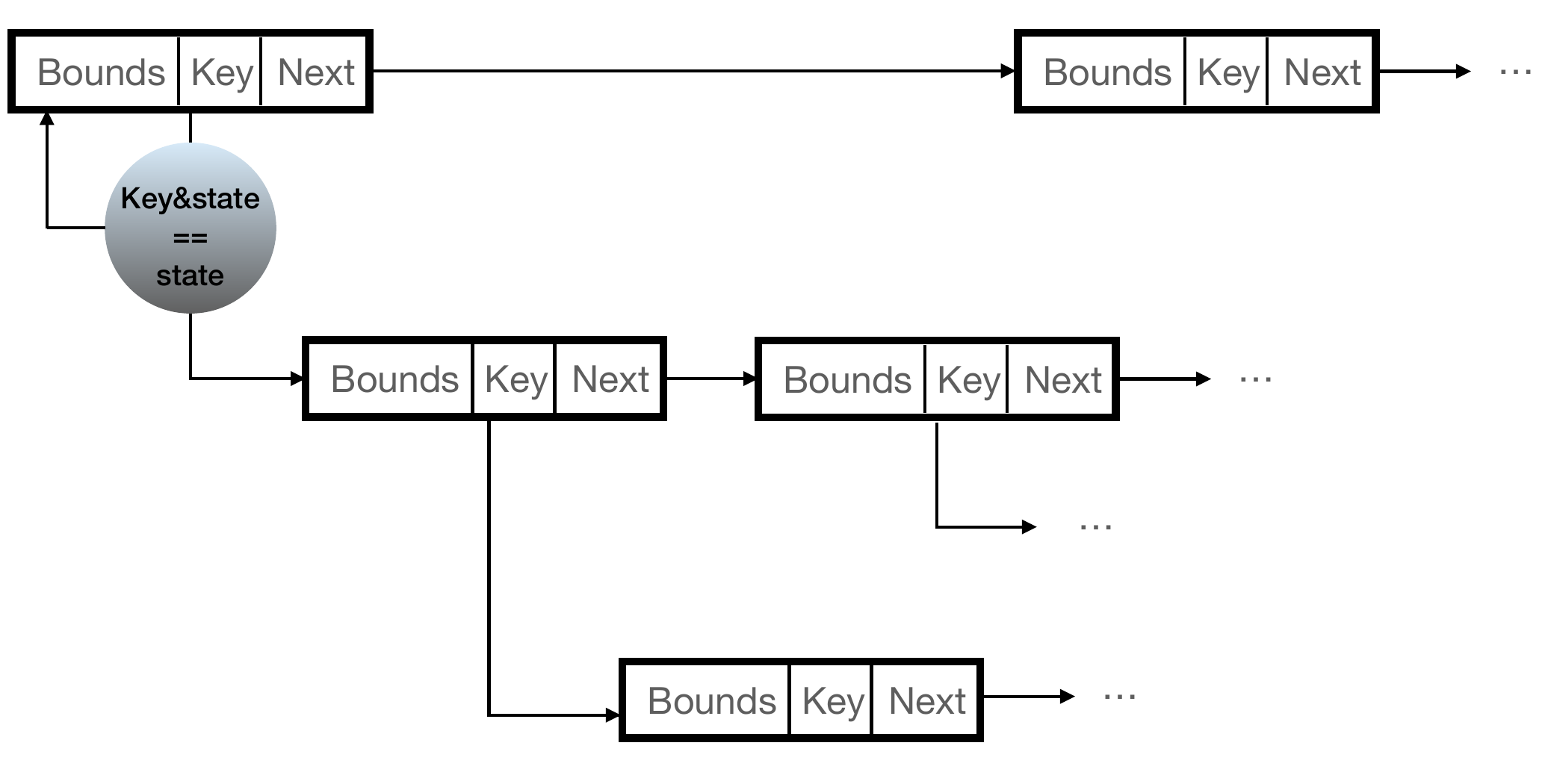}
        \caption{Illustration of a shallow tree structure used to store setrograde databases.  The descent condition for a state lookup is superimposed on the path from the root node to its first child.}
    \label{fig:treeGraph}
\end{figure}

To support fast look-ups and low-cost set operations, our setrograde database uses a specialized shallow tree data structure (implemented as an array, and depicted in Fig. \ref{fig:treeGraph}).  Each node in the database contains:
\begin{itemize}
\item
a 16-bit key, defining positional bits for 4 cards;
\item
an upper and lower bound (4 bits each) on the game value of states in the set (states being defined by the 16-bit keys of all parent nodes and the node itself; cards that are not specified by a parent or the node itself are \emph{x}'s);
\item
and a 32-bit sibling pointer.
\end{itemize}

The maximum tree depth is the same as the retrograde depth, since 4 cards are defined at each level in the tree.  Parents often have many children (the storage cost of each parent is split among its children) with the end effect that for each entry in the 24-card database, there are on average less than 2 (64 bit) nodes.  Additionally, not all leaf nodes are at the maximum tree depth.  \emph{x}'s are implicitly represented, reducing the number of nodes per entry.

An independent tree is stored for each distribution of suits among the four players.  This makes it practical to store the database in a compressed format, and expand only the portions of the database needed at any given point.  Additionally, by reducing average tree size (by a factor of the Lower Bound in Table \ref{fig:stateSpaceSize} — that is a factor of $2\times10^6$ in the case of the 24 card database) we improve locality, and limit the cost of a query.

Several other small optimizations (child pointers are implicit, to save space; sibling pointers are relative) provide memory reductions and improve locality.

% Note that....
% JS: positional bits grows as the number of cards in the deal.
% IS: yes, so the maximum tree depth grows too.  It's a depth-limited tree, but keys high in the tree often have many children.  There are only a few nodes without useful bounds for each node with a useful bound.  With a depth limit of 6, the 24-card database averages less than 2 nodes per entry.  Also there are a lot of \emph{x}'s (which are stored implicitly, saving quite a bit of space.  Only non-x's need to be stored explicitly)

% A node with a 0-offset sibling is terminal.  A node with a 1 offset sibling has no children (a node's first child is always adjacent to the parent).  A node with a sibling pointer value $>1$ has a subtree.

A state can be queried by starting from the root of a tree, and comparing the state's 16-bit key (A state's key has exactly one bit turned on for each 4-bit card representation) with a node's 16-bit key using a bit-wise AND operator.  If the bit-wise AND of the two keys is not equal to the state's key, the state is not represented by the node.  The lookup follows the sibling pointers until a match is found. When a match is found, the bounds are checked, and if a child exists, the search continues with the child.
                
Similar operations are used to look up sets, or given a set to find all dependent sets (sets sharing at least one state).  Bit-wise operations are also used to find sets whose unions can be represented as a single set.

It is likely that the data structure could benefit from additional sorting, allowing for more direct access, however, the benefits of this data structure when compared with other structures we tried were crucial (and sufficient) for generating a 24-card database.

\end{document}